\documentclass[10pt,conference]{IEEEtran}
\usepackage[numbers,sort&compress]{natbib}

\ifCLASSINFOpdf
   \usepackage[pdftex]{graphicx}
   \graphicspath{{figs/}}
   \DeclareGraphicsExtensions{.pdf,.jpeg,.png}
\else
   \usepackage[dvips]{graphicx}
   \graphicspath{{../figs/}}
   \DeclareGraphicsExtensions{.eps}
\fi

\usepackage[cmex10]{amsmath}
\usepackage{amsthm}

\usepackage{algorithmic}

\usepackage{algorithm}
\usepackage{listings}

\floatname{algorithm}{Algorithm}

\lstdefinelanguage{myPascal}[]{Pascal}{
morekeywords={end,bwhile,for,if,else}
}

\usepackage{array}

\ifCLASSOPTIONcompsoc
  \usepackage[caption=false,font=normalsize,labelfont=sf,textfont=sf]{subfig}
\else
  \usepackage[caption=false,font=footnotesize]{subfig}
\fi

\usepackage{url}
\usepackage{multirow}
\usepackage{xcolor}
\newif\iffinal
\finaltrue
\newcommand{\cmtid}{102}

\iffinal
\else
\usepackage[switch]{lineno}
\fi

\begin{document}

\title{ChessMix: Spatial Context Data Augmentation for Remote Sensing Semantic Segmentation}

\iffinal

\author{\IEEEauthorblockN{Matheus Barros Pereira and Jefersson Alex dos Santos}
\IEEEauthorblockA{Department of Computer Science\\Universidade Federal de Minas Gerais, Brazil\\
Belo Horizonte, Minas Gerais, 31270-901\\
Email: \{matheuspereira, jefersson\}@dcc.ufmg.br}}

\fi

\maketitle

\begin{abstract}
Labeling semantic segmentation datasets is a costly and laborious process if compared with tasks like image classification and object detection. This is especially true for remote sensing applications that not only work with extremely high spatial resolution data but also commonly require the knowledge of experts of the area to perform the manual labeling. 
Data augmentation techniques help to improve deep learning models under the circumstance of few and imbalanced labeled samples. In this work, we propose a novel data augmentation method focused on exploring the spatial context of remote sensing semantic segmentation. This method, ChessMix, creates new synthetic images from the existing training set by mixing transformed mini-patches across the dataset in a chessboard-like grid. ChessMix prioritizes patches with more examples of the rarest classes to alleviate the imbalance problems. The results in three diverse well-known remote sensing datasets show that this is a promising approach that helps to improve the networks' performance, working especially well in datasets with few available data. The results also show that ChessMix is capable of improving the segmentation of objects with few labeled pixels when compared to the most common data augmentation methods widely used.
\end{abstract}

\IEEEpeerreviewmaketitle

\section{Introduction} \label{sec:introduction}

Semantic segmentation is the computer vision task whose objective is to classify every pixel from an input image. Many important applications require accurate dense labeling, including remote sensing scenarios, such as for urban monitoring~\cite{resunet}, agriculture~\cite{baeta2017learning} and environmental management~\cite{almeida2015deriving}, as the segmentation provides semantic and localization information cues for interest targets~\cite{pointflow}.

The process of creating labels for a semantic segmentation application is much slower and costly than the amount of work necessary to create labels for classification problems or even object detection ones~\cite{pseudoseg}. This is especially true for remote sensing data, given that in many cases a specialist is required to conduct the labeling process~\cite{song_survey}. Furthermore, acquiring data for remote sensing applications is usually difficult and/or expensive. Aerial images also present their own challenging problems, such as high background complexity, class imbalance, and tiny foreground objects, for instance~\cite{pointflow}.

The aforementioned problems are among the main reasons why there are only a few remote sensing datasets for semantic segmentation publicly available. Furthermore, many of the existing datasets are pretty scarce in terms of annotation quantity and diversity, thus the category types and numbers of labels in them are extremely limited and often fail to meet the data scale requirement of model training~\cite{song_survey}. The consequence is that the performance of semantic segmentation models drops significantly with little pixel-labeled data \cite{pseudoseg}.

Some studies are concerned with data augmentation techniques to address the problem of insufficient or imbalanced training data, by increasing the training sample size and diversity \cite{song_survey}. The most common approaches for data augmentation are geometric and color transformations, random erasing and the addition of synthetic instances to the training set \cite{shorten_survey}. For the latter case, Generative Adversarial Networks (GANs) are a powerful option, but that also imposes many drawbacks, such as the difficulty of converging, the difficulty of generating high-resolution outputs, the requirement of a substantial amount of data to train, and overall high computational cost~\cite{shorten_survey}.

\begin{figure}[!t]
\centering
\includegraphics[width=\linewidth]{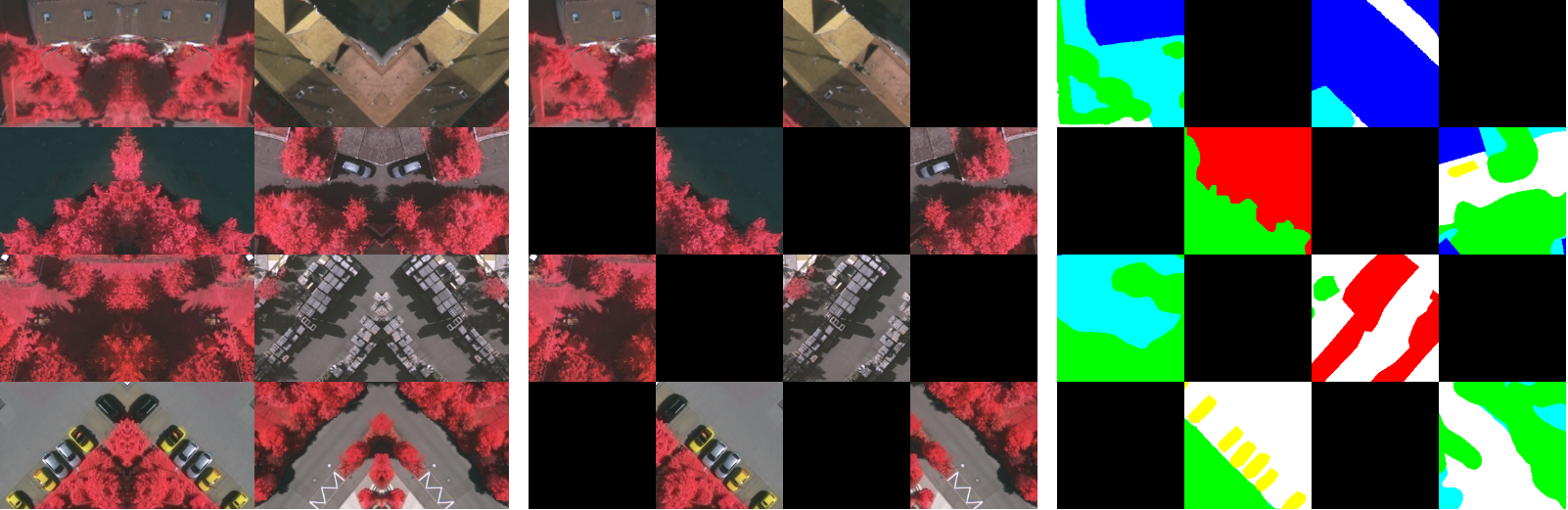}
\caption{Example of a synthetic image generated by the ChessMix data augmentation approach. The leftmost image is the final input image, the rightmost image is the respective thematic map (where black spaces do not propagate the loss), and the middle image is an example of which parts of the input image are effectively learned.}
\label{fig:intro}
\end{figure}

This paper proposes a novel data augmentation method for semantic segmentation tasks on remote sensing images, named ChessMix. Our approach generates new training images by mixing transformed mini-patches from the existing labeled training data. Mini-patches are separated by empty spaces where the loss is not propagated to avoid spatial discontinuation problems. This generates a chessboard-like pattern in the synthetic thematic map. The intuition behind chess organization is to also allow for spatial context relationships between different classes that are not present in the original data. This approach can help to generalize learning about the relationships between pixels of different classes. To mitigate the class imbalance problem, the composing mini-patches are selected so that mini-patches with more examples of the rarest classes are more probable to be selected.

Figure~\ref{fig:intro} illustrates one example of a synthetic image generated with the proposed method. ChessMix unites the upsides from different data augmentation techniques in a single framework, such as the simplicity and versatility of geometric transformations, the robustness to overfitting of random erasing methods, and the capacity of generating new spatial relationships of synthetic image generation methods, while also avoiding the high computational cost of GANs. 

We evaluated the proposed approach on three remote sensing datasets with highly different properties. The results show that the proposed method is capable of improving considerably the segmentation performance under the condition of the low amount of training data and the detection of objects from the rarest classes, while also not compromising the results from the other categories.
\iffinal
 The code for the proposed method is available at \textbf{\url{https://github.com/matheusbarrosp/chessmix}}.
\fi

The remainder of the paper is structured as follows: in Section~\ref{sec:related} we present related works that also employ or propose some type of data augmentation. In Section~\ref{sec:methodology} we detail the proposed data augmentation approach. Section~\ref{sec:setup} presents the datasets that were used to evaluate the effectiveness of the method and the experimental setup. In Section~\ref{sec:results} the semantic segmentation results of ChessMix are presented and discussed. Finally, in Section~\ref{sec:conclusion} we conclude the paper.

\section{Related Work} \label{sec:related}

One of the main motivations for the use of data augmentation is to reduce overfitting and improve the networks' performance \cite{khosla_survey}. However, data augmentation is not the only solution for these problems: regularization (dropout, batch normalization, etc.) and transfer learning, for example, are other types of common techniques that improve the quality of training \cite{shorten_survey}. These solutions are more focused on the characteristics of the model, not the data itself, and can be applied together with the proposed method. Data augmentation is also important to reduce the class imbalance. In this context, some works propose new loss functions (or adaptations to existing ones) to avoid training problems \cite{anantrasirichai2019, hesamian2019, lu2019, bulo2017, liu2020}. These techniques can also be employed in union with ChessMix. Thus, these techniques, although with similar objectives, will not be covered in this paper.

Two broad categories of data augmentation algorithms are usually considered in the literature: data warping and oversampling augmentations \cite{shorten_survey, khosla_survey}. The most common data augmentation methods are from the data warping category, such as geometric transformations (flipping, cropping, rotation, noise injection, etc.). These methods are broadly employed, including in remote sensing applications~\cite{zhao2017, rao2020, stivaktakis2019, liu2018, xue2020, bischke2018}. ChessMix also employs transformations like these, but expands further with the introduction of patch mixing techniques.

Random erasing \cite{zhong2020} is another data warping technique. It removes certain input patches, forcing the model to find other descriptive characteristics \cite{shorten_survey}. Essentially, this characteristic is also present in ChessMix, since the empty spaces between the selected mini-patches from different images also force the model to find different descriptive features. GridMask \cite{gridmask} creates multiple black-spaced regions evenly spaced, while Hide-and-Seek \cite{hideandseek} divides the image into a grid pattern and turns off each grid with an assigned probability. Both of these methods visually resemble our proposed chessboard-like strategy, but in our case, this is done with the additional objective of avoiding spatial inconsistencies between adjacent mini-patches from different images when creating a synthetic one, while random erasing methods only remove parts of the training images (not placing them anywhere).

Regarding the oversampling augmentation methods, which add synthetic instances to the training set, mixing images and GANs are the two most common approaches that work directly in the potential input data \cite{khosla_survey, shorten_survey}. GAN approaches aim at generating realistic synthetic samples for the training \cite{shin2018medical, liu2019gan, ma2019, zheng2019, park2020}. Although carrying a lot of potentials, GANs are known for being highly unstable and difficult to converge \cite{sampath2021}, while also adding a considerable amount of computational cost to the pipeline. Our work differentiates from these cases for not requiring the use of GANs, thus being much lighter and easier to employ.

As for mixing images methods, cut and mix approaches are the closest to the proposed method, since they replace selected regions with some regions of other images \cite{naveed_survey}. CutMix \cite{cutmix} is an augmentation method for image classification that samples images coordinates and replaces the selected patch with a patch from another random image from the mini-batch during training. The Mosaic method from YOLOv4 \cite{yolov4} employs a similar strategy, but instead mixing 4 different images for an object detection task. Methods like these differentiate from ours for two main reasons: \textit{(i)} they do not present a chessboard-like pattern, since they were not created with semantic segmentation problems in mind and, therefore, are not as worried about spatial inconsistencies; \textit{(ii)} the selected patches are not transformed before being inserted into the new image.

Finally, \cite{li2021} employ a class balancing strategy on medical images that adjusts the magnitude of augmentation for different classes by reducing the number of transformed samples for the background classes. ChessMix differentiates by using a different class balancing strategy: our approach gives a probability to every mini-patch that may compose the synthetic images according to the number of pixels of every class. We then select the whole mini-patch based on this probability and perform the transformation in all pixels inside of it.

To the best of our knowledge, this is the first work that studies the use of cut and mix techniques for semantic segmentation in remote sensing scenarios.

\section{Proposed Method} \label{sec:methodology}

\begin{figure}[!t]
\centering
\includegraphics[width=\linewidth]{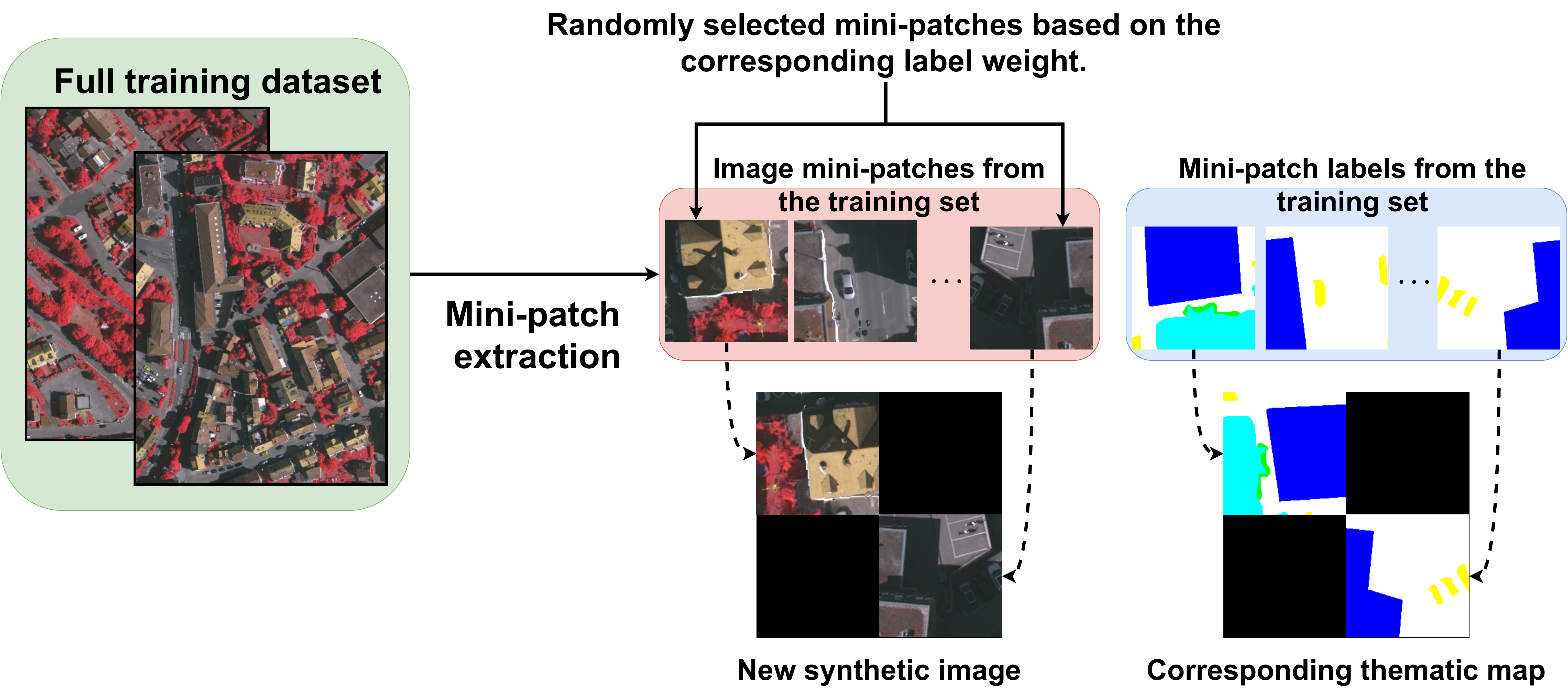}
\caption{Overview of the proposed ChessMix data augmentation strategy.}
\label{fig:methodology1}
\end{figure}

ChessMix is a data augmentation method created especially for the scenario of semantic segmentation with remote sensing data. It can be classified as part of the over-sampling category \cite{shorten_survey}, more specifically under the ``cut and mix''~\cite{naveed_survey} subcategory. ChessMix creates synthetic images by collecting image mini-patches from the different training samples available for the problem and placing them with a transformation in the new image with a chessboard-like grid pattern. This is done in a way that two different mini-patches are not directly on the left, right, above, or below from each other. This is done to avoid spatial discontinuation problems, such as cutting a building in half and placing a tree where the other half should be, while also serving as a way of modeling different spatial context information. Diagonal inconsistencies may still occur, but as the results will later show, this does not compromise the training. The empty (black) grids are also ignored when backpropagating the loss, so that the network will not focus on learning these parts of the images. Figure~\ref{fig:methodology1} shows a visual representation of this process.

The algorithm can be divided into two main steps: mini-patch probability calculation and image generation. The objective of the first step is to pre-calculate the probability of each possible mini-patch (in the training set) being selected to be inserted in the new image. The possible mini-patches are selected as follows. Given a predefined mini-patch size, starting from the first (top-left) pixel, the algorithm selects mini-patches with $50\%$ overlap (both horizontally and vertically), sweeping the image similarly to the kernel of a convolutional network during the convolution process (the stride being half the size of the mini-patch). For each captured mini-patch, a probability weight will be calculated depending on the number of pixels of each class in the mini-patch and stored in a list. The presence of classes with fewer samples carries more weight for the patch. Given the percentage of pixels from each class $c_i$ (calculated from the whole training set), the weight $W_p$ of a mini-patch $p$ is defined as follows:

\begin{equation}
W_p = \sum_{i=1}^{N} (\frac{c_{max}}{c_i} p_i),
\end{equation}

where $N$ is the number of semantic classes of the dataset, $c_{max}$ is the percentage of the class with the most pixels and $p_i$ is the number of pixels from the class $i$ in the mini-patch. For example, given a dataset with 3 classes in which class 1 represents $50\%$ from the whole training dataset, class 2 represents $40\%$ and class 3 represents $10\%$, a mini-patch of size $100\times100$ with 4000 pixels from class 1, 2000 pixels from class 2 and 4000 pixels from class 3 would have the following weight: $W_p = \frac{50}{50}4000 + \frac{50}{40}2000 + \frac{50}{10}4000 = 26500$. Therefore, rarer classes will have higher weights ($c_{max}/c_i$) pondering their number of pixels, causing mini-patches with more examples of less common classes to be more probable of being chosen.

Given the list of pre-calculated mini-patch weights from the previous step, the process of generating new synthetic images can start. The new images will have a predefined size bigger than the mini-patch sizes. In our experiments, we set the size of the images to 2 or 4 times the size of the individual mini-patches, but this value can be adapted according to characteristics of the dataset (such as the overall size of the objects in there). Higher values of mini-patch size will capture bigger objects in the scenes, but will also generate bigger synthetic images, which can potentially be more than the memory of the GPU can handle. To generate a new synthetic image, the algorithm divides it into grids of the same dimension of the mini-patch size, using the chessboard pattern to separate horizontally or vertically adjacent mini-patches. For each one of the valid grid parts, a random mini-patch from the training data is selected to fill the space. The probability of a mini-patch being chosen is proportional to the weight calculated in the previous step. The selected mini-patch is then transformed with data warping augmentation techniques, such as geometric transformations. The semantic segmentation label from the same position of the mini-patch is also transformed in order to follow the changes applied to the mini-patch. It is then inserted in the same grid position of the label being constructed as the input image. Figure~\ref{fig:methodology1} illustrates the whole process. The ChessMix framework is generic enough to allow different types of transformation according to the dataset and problem being worked on. 

Two more options may be considered when generating new images with ChessMix: mirroring mini-patches and mini-patch scales. Mirroring mini-patches are used as not to let black empty grids between the valid mini-patches in the synthetic image (the label will still have them). In this case, each selected mini-patch is mirrored to the immediate left or right empty grid (depending on the row of the grid). This approach is motivated by \cite{resunet}, as the authors affirm that this methodology is particularly useful for aerial images of urban areas due to the high degree of reflecting symmetry these areas have by design.
The mini-patch scales allow the algorithm to generate new images with varying sizes of the grid blocks. For these cases, our approach is to calculate two or more (according to the number of desired scaling factors) lists of mini-patch weight probabilities. After that, when creating a new image, the algorithm first randomly chooses the mini-patch size to be considered, then selects the mini-patches and places them on the grid spaces according to the selection. In a scenario with mini-patch size of $100 \times 100$, new image size of $400 \times 400$ and 2 scales, for example, ChessMix may generate images composed by a $4\times4$ chessboard pattern of grid size $100 \times 100$ or images composed by a $2 \times 2$ chessboard pattern of grid size $200 \times 200$. It is also possible to select varying weights for different scales, allowing images of a certain mini-patch scale to be generated more often than other scales.

The process of creating an image can be then repeated as many times as necessary, in order to generate more training samples for the network. Algorithm~\ref{alg:chessmix_image_generation} describes the whole process.

\begin{algorithm}
\caption{Synthetic image generation by the ChessMix}
\begin{algorithmic}
\STATE $weights \gets calculate\_weights(train\_labels)$
\FOR{each new image}
    \STATE $scale = random(scale\_options)$
    \STATE $new\_img, new\_label = empty(image\_size)$
	\FOR{each valid grid in new\_img}
		\STATE $patch\_index = select\_patch(weights, scale)$
		\STATE $patch = transform(train\_images[patch\_index])$
		\STATE $label = transform(train\_labels[patch\_index])$
		\STATE $new\_img[$valid grid$] = patch$
		\STATE $new\_label[$valid grid$] = label$
		\IF{$mirror\_patch$}
			\STATE $new\_img[$adjacent grid$] = mirror(patch)$
		\ENDIF
	\ENDFOR
	\STATE $save(new\_img, new\_label)$
\ENDFOR 
\end{algorithmic}
\label{alg:chessmix_image_generation}
\end{algorithm}



\section{Experimental Setup} \label{sec:setup}

In this section, we present the ChessMix settings used in the experiments along with the details of the chosen semantic segmentation network. We also present the datasets used to evaluate the proposed approach.

\subsection{Implementation Details}

FCNs\cite{fcn} have been successfully used or adapted for remote sensing applications over the years \cite{sherrah2016, shao2020}. In this work, in order to evaluate the proposed ChessMix method, we selected the FCN-Resnet50 from Pytorch's torchvision models. The FCNs, although effective, are relatively simple (in terms of techniques employed along with the layers) compared to later semantic segmentation methods, such as DeepLab V3\cite{deeplab}. This is the main reason for choosing this type of network to test the ChessMix method, as the reported results will have less bias from the network architecture itself. The weights are initialized from a pre-trained model trained on a subset of COCO train2017, on the 20 categories that are present in the Pascal VOC dataset. For all the following experiments, the network was trained and the best model for the validation set was selected to be used in the test phase. We used the Adam optimizer with learning rate equal to $1e-5$, $0.9$ momentum, and $5e-4$ weight decay. Although we experimented with the FCN-Resnet50 network, the ChessMix method can be used with any semantic segmentation network.

As for the settings of the data augmentation method itself, we generated images in a multiscale approach for each dataset, the scales being 1 or 2. For all datasets, the mini-patch lowest size is one-quarter of the generated image, thus for the scaling factor of 1, the images are composed by a grid pattern of $4 \times 4$ (meaning 8 empty grids and 8 filled ones), while for scaling 2 the pattern is a grid of $2 \times 2$ (2 empty grids and 2 filled ones). We did not give more weight to any of the scaling factors, thus every new image has a $50\%$ chance of being scale 1 or 2.

The transformations applied on the mini-patches were conducted with the use of the Albumentations library \cite{albumentations}. We composed a sequence of transformations as follows:
\begin{enumerate}
    \item Vertical flip with $50\%$ of chance.
    \item Horizontal flip with $50\%$ of chance.
    \item Random rotation by 90 degrees zero or more times with $50\%$ of chance.
    \item Transpose with $50\%$ chance.
    \item Lastly, with $50\%$ chance of occurring, one of the following transformations has an $80\%$ chance of being applied: Grid Distortion or Perspective transformation.
\end{enumerate}

The first 4 transformations are common for many models that use data augmentation as discussed in Section~\ref{sec:related}. The last two (grid distortion and perspective transformation) are used to simulate observed distortions in aerial images \cite{wang2020}. The performed transformations can be altered according to the problem or dataset being worked on, including change of parameters, addition or removal of transformations.

For all datasets, we generated 1000 new images to be added to the original training data using the configuration presented before. 

\subsection{Datasets}

We selected three distinct remote sensing datasets with highly different characteristics to evaluate the proposed ChessMix method. The first is the Vaihingen dataset, provided by the International Society for Photogrammetry and Remote Sensing (ISPRS) Commission for the 2D Semantic Labeling Contest, which contains urban scenes with six different pixel classes. For this dataset, we followed \cite{nogueira2019}, using the areas 11, 15 for validation, areas 28, 30, and 34 for the test, while the remaining areas are used in the training (also used to create the new images with ChessMix). For this dataset, we set the size of the newly generated images to $800 \times 800$ and the size of the mini-patches (grid size) to $200 \times 200$ at scale 1. The training images are also cropped to $800\times800$ size with $50\%$ overlap. The relatively high size of the mini-patches is due to the presence of big objects in the images, such as buildings.

The second dataset is the 2014 IEEE GRSS Data Fusion Contest dataset (Thetford dataset), which is also urban and contains seven pixel classes. For this case, we separated the rightmost small region from the only available test image for the validation, along with the two bottom regions of bare soil present in the middle of the training region. For this dataset, we set the size of the newly generated images to $400 \times 400$ and the size of the mini-patches (grid size) to $100 \times 100$ at scale 1. We also crop the training images with $400\times400$ size and $50\%$ overlap.

Finally, the Brazilian Coffee Scenes dataset (coffee dataset) contains images from four Brazilian cities from the state of Minas Gerais: Arceburgo, Monte Santo, Guaran\'esia, and Guaxup\'e. This is a binary dataset, containing pixels of either coffee or non-coffee crops. Our division protocol for this dataset was using the data from Guaxup\'e as test, Arceburgo as validation, and Guaran\'esia and Monte Santo were used to train the network and perform data augmentation. The size of the new images, mini-patches, and training images' crops follow the same pattern as in the 2014 GRSS Data Fusion Contest dataset.

\section{Results and Discussion} \label{sec:results}

We conducted experiments aiming to answer the following research questions: (1) Can ChessMix improve the semantic segmentation results when compared to the usual data augmentation approach of remote sensing approaches (that is, data warping transformations)? (2) Does ChessMix help to alleviate class imbalance problems and allows better segmentation of the rarest objects?

We evaluate our method with four metrics: overall accuracy (acc), normalized accuracy (norm.acc), intersection over union (IoU, or mean IoU for the non-binary datasets), and Cohen’s kappa coefficient (Kappa). The following subsections present the results and discussion for our research questions.


\subsection{Effectiveness in comparison to the baseline}

For each one of the three datasets, we conducted two experiments (one for the baseline and one for the proposed method). The experiments' settings are as follows:
\begin{enumerate}
    \item Data warping (baseline): the network is trained on the original training set, but applying data warping augmentation during training time. This is the most common data augmentation technique used in the literature, as explained in Section~\ref{sec:related}. To verify how much the ChessMix's chessboard-like spatial distribution truly compares to data warping approaches and to avoid bias due to transformation choice, we use the same transformations employed in the ChessMix process as presented in Section~\ref{sec:setup}. This leaves only our cut and mix strategy as the main difference for a fair comparison. Essentially, at each epoch, the network is seeing a different version of the original training image (the version being a combination of the transformations mentioned in Section~\ref{sec:setup}).
    \item ChessMix+1000: in this approach, we enrich the original training set with new 1000 synthetic images generated by ChessMix.
\end{enumerate}

The networks trained with additional ChessMix images go through more iteration steps per epoch compared to the baseline, as there is more data to be forwarded. Therefore, in order to make fair the comparison between the networks trained with 1000 new synthetic images from ChessMix and the network trained with the baseline (data warping), we employed the following rule: the baseline is allowed to pass through more epochs in order to match the same number of iterations as the network with ChessMix images. In other words, we predefined a number of epochs for the ChessMix+1000 experiments and set the baseline experiment to match the same amount of iterations as the other approach. In the end, both cases took around the same time to complete the training, which means the baseline was not undertrained. The number of epochs for the ChessMix+1000 experiment on the Vaihingen dataset was 300. For the ChessMix+1000 on the Thetford dataset, it was 350 epochs. And finally, for the ChessMix+1000 coffee dataset, it was 400 epochs. Table~\ref{tb:results1} shows the results of the experiments.

\begin{table}[]
\renewcommand{\arraystretch}{1.3}
\caption{ChessMix and baselines results on the three datasets.}
\label{tb:results1}
\centering
\begin{tabular}{llllll}
\hline
\textbf{Dataset}                    & \textbf{Method}       & \textbf{Acc.}  & \textbf{Norm.Acc} & \textbf{IoU}   & \textbf{Kappa} \\
\hline
\multirow{2}{*}{Vaihingen} & Data Warping & 0.848 & 0.688    & 0.601 & 0.798 \\
                           & \textbf{ChessMix+1000}  & 0.856 & 0.697    & \textcolor{red}{\textbf{0.613}} & 0.810 \\
                           \hline
\multirow{2}{*}{Thetford} & Data Warping & 0.935 & 0.860    & 0.809 & 0.897 \\
                           & \textbf{ChessMix+1000}  & 0.913 & 0.943    & \textcolor{red}{\textbf{0.836}} & 0.873 \\
                           \hline
\multirow{2}{*}{Coffee}    & Data Warping & 0.914 & 0.847    & \textcolor{red}{\textbf{0.616}} & 0.710 \\
                           & \textbf{ChessMix+1000}  & 0.915 & 0.846    & \textcolor{red}{\textbf{0.616}} & 0.711
                           \\
                           \hline
\end{tabular}
\end{table}

As we can see in Table~\ref{tb:results1}, adding 1000 synthetic images from the proposed ChessMix method to the original dataset made the IoU achieve the highest value for two of the evaluated datasets and was tied with the baseline for the Coffee dataset. The amount of improvement highly depends on the properties of the dataset. For example, the Thetford dataset is the one with the least number of training images, the least number of labeled pixels for the training, and also the one with the most number of classes. These properties make it harder for the network to accurately learn how to segment the objects of the dataset. In this type of situation, data augmentation helps to generate more samples and even reduces the class imbalance problem (in the case of ChessMix). This is the reason why the improvement brought by ChessMix when compared to the baseline is the highest in this dataset. Applying data warping transformations, the mean IoU was $0.80$ in the Thetford dataset, while with the use of ChessMix samples these results were further improved, achieving $0.83$ mean IoU. This big difference to the data warping baseline can also be seen in the normalized accuracy, as it was increased from $86\%$ to $94\%$, showing that ChessMix is more accurate when predicting pixels from the rarest classes. The baseline for this case achieved $2\%$ higher overall accuracy and Kappa results due to more correct predictions for the second most common class (road). The drawback was the higher mislabeling frequency of the classes with fewer examples. This is confirmed by the results shown in Figure~\ref{fig:thetford1}.

A similar improvement is not observed in the other two datasets. There are many reasons for this. First, they are relatively rich in terms of training data, which allows even the baseline to achieve high results. But even under this condition, ChessMix proved to be able to increase the mean IoU in the Vaihingen dataset. For this dataset, there was a specially important improvement in the car label (the class with fewer pixels apart from background/clutter for this dataset), which will be further discussed in the next research question subsection.

As for the coffee dataset, the main reasons for the lack of improvement are the number of classes of the dataset (only 2), and the high amount of intraclass variance, which hinders the learning of the network even with the presence of complex data augmentation techniques. But even for this case, the results are so close that it can be considered a draw between the two experiments.

\subsection{Few labeled data evaluation}

\begin{figure*}[!t]
\centering
\subfloat[Data Warping]{\includegraphics[width=3.2in]{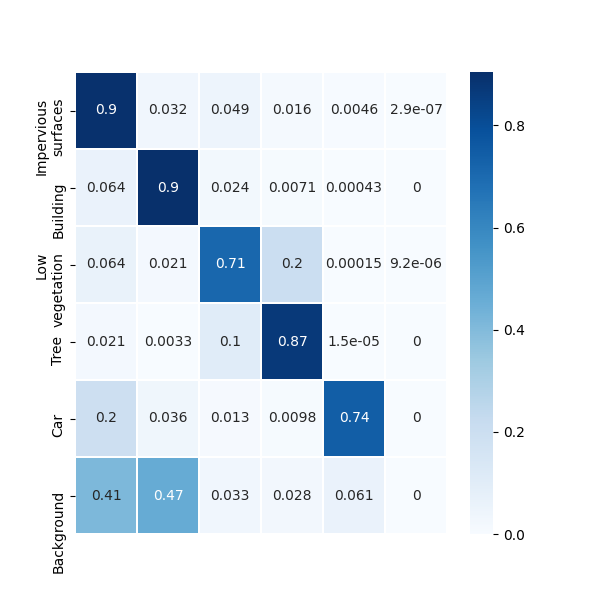}%
\label{fig:vaihingen1}}
\hfil
\subfloat[+1000 ChessMix samples]{\includegraphics[width=3.2in]{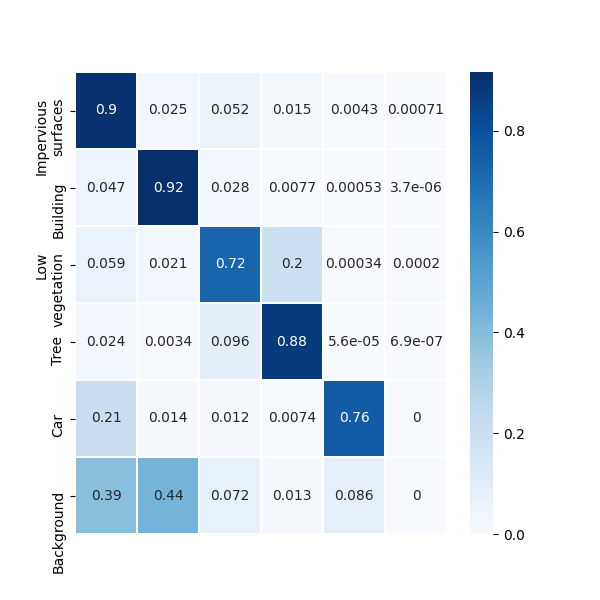}%
\label{fig:vaihingen2}}
\caption{Class accuracies of the baseline data warping method and by adding 1000 samples from ChessMix for the Vaihingen Dataset.}
\label{fig:vaihingen}
\end{figure*}

\begin{figure*}[!t]
\centering
\subfloat[Data Warping]{\includegraphics[width=3.2in]{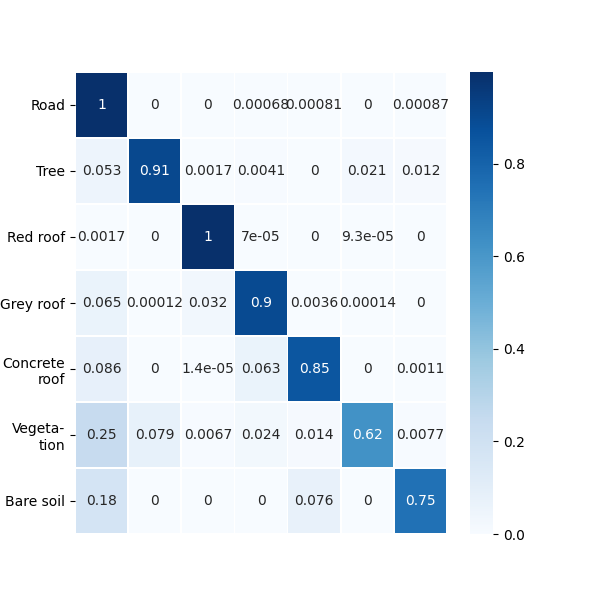}%
\label{fig:thetford1}}
\hfil
\subfloat[+1000 ChessMix samples]{\includegraphics[width=3.2in]{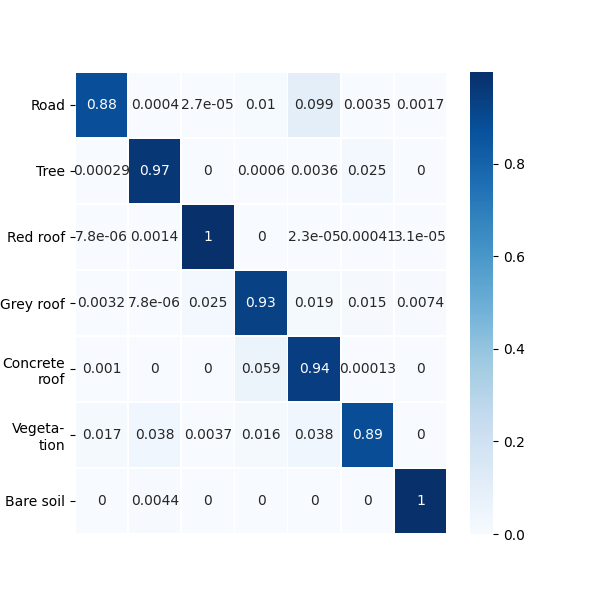}%
\label{fig:thetford2}}
\caption{Class accuracies of the baseline data warping method and by adding 1000 samples from ChessMix for the Thetford Dataset.}
\label{fig:thetford}
\end{figure*}

In order to evaluate how well ChessMix performed on the labels with few examples, we must analyze the class individual results. Figures~\ref{fig:vaihingen} and \ref{fig:thetford} show the class accuracy of both the cases of applying data warping and adding 1000 ChessMix samples for the Vaihingen and Thetford datasets.

By looking at the Vaihingen results in Figure~\ref{fig:vaihingen}, we can see that the use of ChessMix images improved the accuracy of cars from $74$ to $76\%$. This shows how the proposed method can help alleviate the imbalance problem better than the most commonly used data augmentation methods. Both methods still mislabel the background/clutter pixels, but considering this is the class with fewer pixels and with extremely high variance, as it contains pretty much everything that is not one of the other classes, this is an expected result. We can also see an improvement for the class "Building", which was also increased by $2\%$.

The most notable improvements are, however, in the Thetford dataset, as seen in Figure~\ref{fig:thetford}. For this case, the use of ChessMix images greatly improved the accuracy of the class bare soil, which went from $75\%$ in the baseline to almost $100\%$ with ChessMix augmentation. Concrete roof and vegetation were also considerably improved: the first went from $85\%$ to $94\%$ and the latter went from $62\%$ to $89\%$. Smaller, but still considerable, improvements are also present for the ``Grey roof" ($3\%$) and ``Tree" ($6\%$) classes. For this dataset, the baseline only won (by $12\%$) considering the accuracy of the road class, which is the second most common in the dataset. In this context, we can also see the amount of mislabeled road pixels by the baseline, mostly for vegetation and bare soil. The difference in the normalized accuracy score between the two cases ($86\%$ for the data warping baseline and $94\%$ for the ChessMix) shows, however, that although losing in the road class, ChessMix was able to balance more the accuracy of the classes with fewer examples.

\section{Conclusion} \label{sec:conclusion}

In this paper, we proposed a novel data augmentation strategy called ChessMix, which takes advantage of the different spatial context information from remote sensing semantic segmentation datasets. The method mixes transformed mini-patches from all the labeled training set of a dataset in a chessboard grid pattern. This strategy not only allows the existence of different spatial dependencies than the original images, but also avoids adjacent discontinuation problems by not back-propagating the loss from the ``empty'' grids.

We evaluated ChessMix with three highly different remote sensing datasets and compared its results to the data warping approaches usually employed in the literature. The results show that ChessMix is a promising method, capable of improving the performance of a semantic segmentation network when compared to the use of data warping techniques, especially in situations of few labeled data. Furthermore, ChessMix proved to be a reliable option to improve the accuracy of classes with few labeled examples, such as the bare soil class from the Thetford dataset and cars in the Vaihingen dataset.

For future works, we plan on further exploring ChessMix through a deeper ablation study. More refined techniques can also be considered for the process of placing transformed mini-patches in the synthetic images, such as trying to make common specific types of rare spatial dependencies in the original training data.

\section*{Acknowledgment}

This work was supported by the Serrapilheira
Institute (grant number Serra – R-2011-37776). 
This work was supported in part by the Minas Gerais Research Funding Foundation (FAPEMIG) under Grant APQ-00449-17, by the National Council for Scientific and Technological Development (CNPq) under Grant 311395/2018-0 and
Grant 424700/2018-2, and  by the \emph{Coordenação de Aperfeiçoamento de Pessoal de Nível Superior – Brasil (CAPES)} – Finance Code 001. We gratefully acknowledge the support of NVIDIA Corporation with the donation of the Titan Xp GPU used for this research.

\bibliographystyle{IEEEtran}

\end{document}